%% file: pricai506.tex
\documentclass[runningheads]{llncs}
\usepackage{amsmath,amsfonts}

\usepackage{array}
\usepackage{textcomp}
\usepackage{stfloats}
\usepackage{url}
\usepackage{verbatim}
\usepackage{graphicx}
\usepackage{cite}
\usepackage{algorithm,algpseudocode,graphicx,float,multirow}
\usepackage{enumitem}
\usepackage{color}
\usepackage{CJKutf8}
\usepackage{pifont}
\usepackage{setspace}
\usepackage[misc]{ifsym}
\usepackage{hyperref}

\usepackage[T1]{fontenc}
\usepackage{graphicx}
\usepackage{color}

\begin{document}

\title{UTBoost: Gradient Boosted Decision Trees for Uplift Modeling}

\author{
    Junjie Gao\textsuperscript{\Letter} \and Xiangyu Zheng \and DongDong Wang \and Zhixiang Huang \and \\Bangqi Zheng \and Kai Yang
}

\authorrunning{Junjie et al.}

\institute{JD Technology, Beijing, China\\
\email{\{gaojunjie10,zhengxiangyu8,wangdongdong9,huangzhixiang,\\zhengbangqi,yangkai188\}@jd.com}}

\maketitle
\vspace{-0.5cm}

\begin{abstract}
    Uplift modeling comprises a collection of machine learning techniques designed for managers to predict the incremental impact of specific actions on customer outcomes. 
    However, accurately estimating this incremental impact poses significant challenges due to the necessity of determining the difference between two mutually exclusive outcomes for each individual.  
    In our study, we introduce two novel modifications to the established Gradient Boosting Decision Trees (GBDT) technique.
    These modifications sequentially learn the causal effect, addressing the counterfactual dilemma. 
    Each modification innovates upon the existing technique in terms of the ensemble learning method and the learning objective, respectively. 
    Experiments with large-scale datasets validate the effectiveness of our methods, consistently achieving substantial improvements over baseline models.
\end{abstract}

\input{content/introduction.tex}
\input{content/methodology.tex}
\input{content/tddp.tex}
\input{content/causal_gbm.tex}
\input{content/experimental_evaluation.tex}
\input{content/conclusion.tex}

\bibliographystyle{splncs04}
\bibliography{./ref.bib}

\end{document}

%% file: content/introduction.tex
\section{Introduction}
Uplift modeling, a machine learning technique used to estimate the net effect of a particular action, has recently drawn considerable attention. 
In contrast to traditional supervised learning, uplift models concentrate on modeling the effect of a treatment on individual outcomes and generate uplift scores that show the probability of persuasion for each instance. 
This technique has proven particularly useful in personalized medicine, performance marketing, and social science. 

However, a notable challenge in uplift modeling is the absence of observations for both treated and control conditions for an individual within the same context. 
Various uplift tree-based studies have tackled this challenge by creating measures of outcome differences between treated and untreated observations and maximizing heterogeneity between groups \cite{rzepakowski2012decision,2016Recursive}. 
Several research studies provide further generalization to bagging ensemble methods on the idea of random forests \cite{guelman2012random, guelman2015uplift}, which aim to address the challenges of tree model performance decay with an increasing number of covariates. 
Despite the success of these nonparametric methods in prediction, we have experimentally discovered that random-forest based methods still suffer from significant degradation of power in predicting causal effects as the number of variables increases.

Boosted trees, with their iterative refinement capabilities, offer a promising solution. 
By sequentially fitting trees to residuals from previous models, boosting methods can effectively capture intricate patterns in the data, leading to more accurate uplift estimates.
In this paper, we propose an innovative approach to the uplift tree method, which employs gradient boosting as the ensemble technique.  
Subsequent trees are fitted based on the causal effect learned by preceding models. 
Our findings indicate that while both ensemble learning methods perform comparably in low-dimensional settings, boosting exhibits significant advantages in handling high-dimensional data.

However, the split criterion employed in uplift tree fails to consider the learning of outcomes and instead focuses solely on uplift signals. 
An alternative approach is to utilise the outcomes observed under different treatments as a fitting objective, with meta-learning, in particular, gaining traction for its versatility in employing any machine learning estimator as a base model.
The Single-Model \cite{athey2015machine} and Two-Model \cite{radcliffe2007using} approaches represent two prevalent strategies.
The first method is trained over an entire space, with the treatment indicator serving as an additive input feature. 
However, this method's drawback is that a model may not select the treatment indicator if it only uses a subset of features for predictions, such as a tree model. 
Consequently, the causal effect is estimated as zero for all subjects. 
An enhancement over the Single-Model method involves using two separate models to represent the two potential outcomes. 
However, this dual-model approach does not leverage the shared information between control and treated subjects, resulting in cumulative errors \cite{radcliffe2011real}.
Furthermore, additional model training also incurs higher training resource consumption and model deployment costs.

In this paper, we propose CausalGBM (Causal Gradient Boosting Machine), a new nonparametric method that utilizes tree models as base learners to simultaneously learn causal effects and potential outcomes through loss optimization.
Our method explicitly computes the contribution of causal effects to the loss function at each split selection, avoiding the defect that the treatment indicator may not be picked up in some scenarios where the causal effects are weak. 
We further enhance the model's convergence speed by incorporating second-order gradient information.
Experimental comparisons on four datasets demonstrate that our model outperforms baseline methods and shows better robustness.
We have implemented our novel techniques in the UTBoost (Uplift Tree Boosting system) software\footnote{\href{https://github.com/jd-opensource/UTBoost}{https://github.com/jd-opensource/UTBoost}}, available under the MIT license. 
We highlight our contributions as follows:
\begin{enumerate}[topsep=0pt]
	\item We innovate the uplift tree method that focuses on maximizing heterogeneity by extending the ensemble of trees from bagging to boosting.
	\item For the first time, we integrate potential outcomes and causal effects within the classical GBDT framework, employing a second-order method to fit the multi-objective function. 
In the context of randomized trials, we propose an approximation method that significantly reduces the computational complexity of the algorithm.
	\item Through extensive experimentation on real-world and public datasets, we demonstrate that our models outperform baseline methods and exhibit superior robustness.
\end{enumerate}

%% file: content/methodology.tex
\section{Uplift Problem Formulation}
Uplift modeling seeks to quantify the incremental causal effect of interventions on individual outcomes. 
Given $n$ independent and identically distributed samples $\{(\mathbf{X}_i,y_i,w_i)\}_{i=1}^{n}$, where each sample comprises $p$ features $\mathbf{X}_i \in \mathbb{R}^p$, an observed outcome $y_i \in \mathbb{R}$, and a binary treatment indicator $w_i$, which indicates whether unit $i$ received treatment ($w_i = 1$) or control ($w_i = 0$). 
Let $y_i(1)$ and $y_{i}(0)$ be the potential outcomes that would be observed for unit $i$ when $w_i=1$ and $0$, respectively.
The uplift of an individual $i$, denoted by $\tau_i$, is denoted as: $\tau_i=y_i(1)-y_i(0)$.

In practice, we will never observe both $y_i(1)$ and $y_{i}(0)$ for a same individual and thus $\tau_i$ cannot be directly calculated.
Fortunately, we can use the conditional average treatment effect (CATE) as an estimator for the uplift.
In the uplift modeling literature \cite{2021A}, it is typical to assume that the treatment $w_i$ is randomly assigned and CATE is defined as:
\begin{equation}
	\tau(\mathbf{x})=
	\mathbb{E}[y|w\!=\!1, \mathbf{X}\!=\!\mathbf{x}]
	-
	\mathbb{E}[y|w\!=\!0, \mathbf{X}\!=\!\mathbf{x}],
	\label{eq.def1}
\end{equation}
which signifies the uplift on $y$ caused by the treatment $w$ for the subject with feature $\mathbf{x}$.
Uplift can be empirically estimated by considering two groups: a treatment group and a control group (without treatment).

%% file: content/tddp.tex
\section{Tree Boosting for Treatment Effect Estimation}
\label{sec.boost}

Our first proposed method adopts a sequential learning approach to fit uplift directly.
This method extends the ensemble learning approach of uplift tree, and it has better performance on high-dimensional data compared to the uplift random forest method. 
As the splitting criterion in the training process is similar to the standard delta-delta-p (DDP) algorithm \cite{hansotia2002incremental}, which aims to maximize the difference of uplift between the left and right child nodes. 
We refer to this method as TDDP (Transformed DDP). 
It enables incremental training of subsequent uplift trees by transforming the sample labels in each iteration.

\subsection{Ensemble Learning with Transformed Labels}
Taking the decision tree as the base learner, we use a sequence of decision trees to predict the uplift $\tau(\mathbf{x})$. 
As the uplift cannot be observed for each sample unit, the ensemble method for $\tau(\mathbf{x})$ differs from the common supervised-learning scenarios.
We explicitly derive the optimization target for uplift estimation in each iteration of the boosting.

Denoting a tree model by $T(\mathbf{x};\theta_j)$, where $\theta_j$ encapsulates the model's parameters, including partitioning and leaf node estimation, our goal is to predict the uplift through the equation:
$
	\widehat{\tau}(\mathbf{x})
	=
	\sum_{j=1}^{M}T(\mathbf{x}; \theta_j).
$
We sequentially optimize $T(\mathbf{x}; \theta_j)$ to minimize the loss associated with $\widehat{\tau}(\mathbf{x})$.

Let $u_{m}=\sum_{j=1}^{m}T(\mathbf{x}; \theta_j)$ represent the cumulative prediction of the first $m$ trees. 
At step $m$, with the current prediction $u_{m}$, the loss is denoted by ${\mathcal{L}(\tau(\mathbf{x}), u_{m}(\mathbf{x}))}$. 
The gradient, then, is defined as 
$
	\boldsymbol{g}_m:=\frac{\partial
		\mathcal{L}(\tau(\mathbf{x}), u_m(\mathbf{x}))
	}
	{
		\partial{u_m}(\mathbf{x})
	}
$
, where $-\boldsymbol{g}_m$ indicates a local direction for further decreasing the loss at $u_m$. 
Thus, in a greedy manner, we fit $T(\mathbf{x};\theta_{m+1})$ to approximate $\boldsymbol{g}_m$.
Specifically, for uplift modeling, the quadratic loss at step $m$ can be expressed as:
\begin{equation*}
	\begin{aligned}
		\mathcal{L}(\tau(\mathbf{x}), u_m(\mathbf{x}))
		= 
		\frac{1}{2}
		\Big\{\mathbb{E}[y|\mathbf{X}=\mathbf{x},w=1]-\mathbb{E}[y|\mathbf{X}=\mathbf{x},w=0]
		-u_m(\mathbf{x})\Big\}^2
	\end{aligned}
\end{equation*}
with the coefficient $\frac{1}{2}$ simplifying gradient computation. 
The negative gradient then becomes:
\begin{equation*}
	\begin{aligned}
		-
		\frac{
			\partial\mathcal{L}(\tau(\mathbf{x}), u_m(\mathbf{x}))
		}{
			\partial u_m(\mathbf{x})
		} 
		&=
		\mathbb{E}[y-u_m(\mathbf{x})|\mathbf{X}=\mathbf{x},w=1]
		-
		\mathbb{E}[y|\mathbf{X}=\mathbf{x},w=0].
	\end{aligned}
\end{equation*}
Thus, in constructing the $(m+1)$-th tree, $T(\mathbf{x};\theta_{m+1})$, we transform the outcome $y_i$ to $y_i - u_m(\mathbf{X}_i)$ for treated units, while maintaining the original outcome for the control group. 
This approach allows for tree construction based on these transformed outcomes. 
Algorithm \ref{alg.boosting} outlines the overall training procedures for TDDP, where the \emph{split criterion} temporarily serves as a placeholder and the details will be introduced in the following subsection \ref{sec.tree_construct}.

\begin{algorithm}\caption{Gradient Tree Boosting for Uplift Modeling}
	\begin{algorithmic}[1]
		\Require Data: $\mathcal{D} = \left\{(\mathbf{X}_i, y_i, w_i)\right\}_{i=1}^{N}$, Shrinkage rate: $\alpha$
		\Ensure $u_M=\sum_{m=1}^{M}T(\mathbf{x};\theta_m)$
		
		\State Set $u_0(\mathbf{x})= 0$.
		\For{$m=1,\cdots ,M$} 
		
		\State 
		Set $y_i=y_i-T(\mathbf{x}, \theta_{m-1})$ for $\{i~|~w_i=1\}$.
		
		\State{\textbf{Build Tree Structure:} Recursively partition $\mathcal{D}^m$:}
		\indent \While{the \emph{stopping rule} is not satisfied}
		\State{Select the optimal split $s^*$ in the candidate splits by criterion \eqref{criterion}.}
		\State{Split the current node into child nodes by $s^*$.}
		\EndWhile
		\State{Output the Tree Structure $T_m$}
		
		
		\State{\textbf{Obtain the Estimator $T(\mathbf{x};\theta_m)$:}
		}
		
		\For{each leaf node $t_i$ of $T_m$}
		
		\State{Get $D^{m}(t_i)$:
			the sample units in $D^{m}$ that fall into $t_i$.
		}
		\State{Estimate Weight: $\widehat{\tau}_m(t_i)=\bar{Y}_1(D^{m}(t_i))-\bar{Y}_0(D^{m}(t_i))$.
		}
		
		
		\State{Output the m-th predictor:
			$T(\mathbf{x}; \theta_m) = \alpha\widehat{\tau}_m(t_{T_m}(\mathbf{x}))$, where 
			$t_{T_m}(x)$ denotes the leaf node that $x$ belongs to in $T_m$.
		}
		\EndFor
		\EndFor
	\end{algorithmic}
	\label{alg.boosting}
\end{algorithm}

\subsection{Tree Construction Method}
\label{sec.tree_construct}
Here, we delve into the split criterion, a pivotal component of our tree construction methodology.

\paragraph{\textbf{Split Criterion}}
Traditional CART algorithms select splits to minimize mean squared errors (MSE) in regression trees. 
However, this approach is not directly applicable in uplift modeling due to the unobservability of unit-level uplift ($\tau_i$). 
In our context, we adapt this criterion to leverage aggregated uplift statistics, such as averages or variances, available within groups of units.

In the next, we will show that minimizing the MSE is equivalent to maximizing the gaps between the average uplift within the split nodes.
Consider the split selection at an internal root node $t$ with data $D_t:=\{\mathbf{X}_i, y_i, w_i\}_{i=1}^{n_t}$. 
Let $s$ denote a split, $s_L$ and $s_R$ denote the indices set in the left and right child nodes with sub-sample size $n_L$ and $n_R$, respectively, under the split ${s}.$
For example, suppose $s = \{x_j = a\}$ for a numeric variable $x_j$, then $s_L=\{i|X_{ij}\leq a\}$ and $s_R=\{i|X_{ij}> a\}$. 
Let $\bar{\tau}_{L}:=\sum_{i\in s_{L}}\frac{\tau_i}{n_{L}}$ and $\bar{\tau}_{R}:=\sum_{i\in s_{R}}\frac{\tau_i}{n_{R}}$ denote the average uplift in the left and right child nodes, respectively.
Then we have the following proposition: 

\begin{proposition}
	Minimizing the mean squared errors of $\tau_i$ in the split nodes is equivalent to maximizing the difference between the average uplift within the left and right child nodes, \textit{i.e.},
	\begin{equation*}
		\begin{aligned}
			\mathop{arg}\!\mathop{min}_s
			\Big\{ \!\!
			\sum_{i|i\in s_L}\!\!\big(\tau_i -\bar{\tau}_L\big)^2
			\!\!+\!\!\!\!
			\sum_{i|i\in s_R}\big(\tau_i -\bar{\tau}_R\big)^2
			\Big\} 
			\!
			\!=\!
			\mathop{arg}\!\mathop{max}_s
			\big\{ 
			\frac{n_Ln_R}{n}
			(\bar{\tau}_L-\bar{\tau}_R)^2
			\big\}.
		\end{aligned}
	\end{equation*}
	\label{prop.1}
	\vspace{-0.25cm}
\end{proposition}
\noindent Proposition \ref{prop.1} guides us to a practical split criterion for the uplift modeling as both $\bar{\tau}_{L}$ and $\bar{\tau}_{R}$ are aggregated values that can be estimated from data.
Taking $\bar{\tau}_{L}$ as an example, the definition of $\bar{\tau}_L$ involves $\{y_i(1), y_i(0)\}$ as shown in equation \eqref{eq.tau_l},
\begin{equation}
	\bar{\tau}_{L}
	=
	\frac{\sum_{i\in s_L}y_i(1)-y_i(0)}{n_L}
	=
	\bar{Y}_{L}(1)-\bar{Y}_L(0).
	\label{eq.tau_l}
\end{equation}
Under randomly assigned treatment, $\bar{Y}(1)$ and $\bar{Y}(0)$ can be estimated by the sample average of $y$ in the treatment and control groups, respectively.
Therefore, the optimal split $s^{*}$ is selected by the following rule:
\begin{equation}
	s^{*}=\arg\max_{s}
	\Big\{ \!\!
	\frac{n_Ln_R}{n}
	\big[
	\big(\bar{Y}_{L}^{1}-\bar{Y}_{L}^{0}\big)
	- 
	\big(\bar{Y}_{R}^{1}-\bar{Y}_{R}^{0}\big)
	\big]^2
	\Big\},
	\tag{\textit{split criterion}}
	\label{criterion}
\end{equation}
where $Y_{L}^{1}:=\frac{\sum_{i|i\in s_L, w_i=1}y_i}{n_L^1}$ with $n_{L}^1$ denoting the number of treated units in the left child node, and $Y_L^0$, $Y_R^1$, $Y_R^0$ are defined similarly.

It's important to note that TDDP, while inspired by gradient boosting, diverges from it by focusing on observed outcomes and directing weak learners towards uncovering heterogeneities in treatment effects rather than strictly following a gradient descent path.
Moreover, as TDDP is exclusively concerned with uplift as the learning objective, this method is unsuitable for estimating potential outcomes.
In the following section, we put forward an alternative gradient boosting method which takes both causal effects and potential outcomes as dual learning objectives and adheres strictly to the gradient descent path.


%% file: content/causal_gbm.tex
\section{Causal Gradient Boosting Machine}

Single-Model and Two-Model are two widely used strategies. However, Single-Model suffers from model invalidation due to unselected treatment indicator variables, while Two-Model's dual model incurs accumulated errors \cite{radcliffe2011real} and doubled training and deployment costs.
In order to alleviate the shortcomings of the two approaches, we propose a Causal Gradient Boosting Machine (CausalGBM) to fit causal effects and outcomes in a single learner.
This approach extends the standard gradient boosting algorithm to the field of causal effect estimation, thus bridging the gap between the two classes of methods.
Since this technique calculates the contribution of causal effects to the loss separately at each split, it does not suffer from the problem of model failure that may occur in Single-Model, and compared with the Two-Model method, it can train a single model using the information of the whole samples set, avoiding the accumulation of errors by multiple models.

\subsection{Learning Objective}
In order to realize the simultaneous estimation of the two objectives, we split the original single learning task.
We can conduct that:
\begin{equation}
	y_i=y_i(1)w_i+y_i(0)(1-w_i)=w_i\tau_i+y_i(0)
	\label{eq.obj}
\end{equation}
which indicates that for treated instances, observed outcomes are the sum of potential outcomes and individual causal effects, while for control instances, they equate to the potential outcomes alone. 
This relationship facilitates the indirect learning of both potential outcomes and individual causal effects from the observed data.

We employ a tree ensemble model with $2M$ additive functions to predict the output for a dataset with $n$ samples and $p$ features:
$$
\hat{y_i}=\sum_{m=1}^{M}f_m(\mathbf{X}_i)+w_i\tau_m(\mathbf{X}_i),\ f_m,\tau_m\in\mathcal{F}
$$
where $\mathcal{F}=\{f(\mathbf{X})=v_{q(\mathbf{X})},\tau(\mathbf{X})=u_{q(\mathbf{X})}\}(q:\mathbb{R}^p \rightarrow T,v\in \mathbb{R}^T,u\in \mathbb{R}^T)$ represents the regression trees space. 
Here $q$ maps each example to the corresponding leaf index in each tree, and $T$ refers to the number of leaves in the tree. 
Note that leaf weights comprise both $u$ and $v$ in this framework, which is significantly different from the classical regression tree. 
We will use the decision rules in the trees (given by $q$) to classify instances to leaves and compute the final predictions by summing up the scores by \eqref{eq.obj} in the corresponding leaves (given by $u, v$). 
To learn the set of functions that are employed in the ensemble model, we minimize the following objective function:
$
\mathcal{L}(\theta)=\sum_il(y_i,\hat{y_i})
$.
Here $l$ is a differentiable convex loss function that measures the difference between the prediction and the observed label. Using the binary decision tree as a meta-learner, we train the ensemble model sequentially to minimize loss. In other words, let $\hat{y_i}^t$ be the prediction of the i-th instance at the t-th iteration, we add $f_t+w_i\tau_t$ to minimize the following objective:
$$
\begin{aligned}
\mathcal{L}^{(t)}
	&=\sum_{i=1}^n l(y_i,\hat{y_i}(0)^{(t-1)}+f_t(\mathbf{X}_i)+w_i(\hat{\tau_i}^{(t-1)}+\tau_t(\mathbf{X}_i)))
\end{aligned}
$$
We employ a second-order approximation to expedite the optimization process.

Under the setting that $w_i \in [0, 1]$, we can remove the constant terms to obtain the following simplified objective at step $t$: 
$$
\begin{aligned}
\tilde{\mathcal{L}}^{(t)}
 =\sum_{i=1}^n [
w_i g_i \tau_t(\mathbf{X}_i) +
\frac{1}{2} w_i h_i \tau_t^2(\mathbf{X}_i)
+
g_i f_t(\mathbf{X}_i)
 +
\frac{1}{2} h_i f_t^2(\mathbf{X}_i) +
w_i h_i \tau_t(\mathbf{X}_i)f_t(\mathbf{X}_i)
]
\end{aligned}
$$
where $g_i=\partial_{\hat{y}^{(t-1)}}l(y_i,\hat{y}^{(t-1)})$ and $h_i=\partial_{\hat{y}^{(t-1)}}^2l(y_i,\hat{y}^{(t-1)})$ are first and second order gradient statistics on the loss function. 
Note that they are defined in the same way as standard gradient trees.

Define $I_j = \{i|q(\mathbf{X}_i) = j\}$ as the instance set of leaf $j$, we can rewrite the above equation as:
$$
\begin{aligned}
\tilde{\mathcal{L}}^{(t)}&=\sum_{j=1}^T [
(\sum_{i \in I_j} w_ig_i + w_ih_if_j)\tau_j + \frac{1}{2}(\sum_{i \in I_j}w_ih_i)\tau_j^2 
+(\sum_{i \in I_j} g_i)f_j + \frac{1}{2}(\sum_{i \in I_j}h_i)f_j^2
]
\end{aligned}
$$
We can further derive the optimal values for $f_j$ and $\tau_j$ of this dual quadratic function and the corresponding optimal weights $v^*$ and $u^*$. 
However, it is important to note that solving for both weights simultaneously will result in a significant decrease in the computing efficiency of the algorithm, compared to the standard regression tree, which has a simpler analytic solution for the quadratic function, during training process. 
We will introduce an approximation method to solve this difficulty in the next section.

\subsection{Multi-objective Approximation}
We point that if all $w_i$ are equal to $0$, i.e., the data set contains only control samples, the above equation is identical to the optimization objective of the regression tree, and we can compute the optimal $v_0^*$ in that specific context. 
Under the setting that treatments are assigned randomly, we further assume that $v^*=v_0^*$ on each leaf, which enables us to derive the optimal weights $v^*$ with control instances. 
After that, the objective function degenerates to a simple quadratic function with one variable and we can solve optimal $u^*$. We can compute the optimal weights by
$$
\begin{aligned}
v_j^*=-\frac{\sum_{i \in I_j^0} g_i}{\sum_{i \in I_j^0} h_i}, \quad  
u_j^*=-\frac{\sum_{i \in I_j} w_ig_i+w_ih_iv_j^*}{\sum_{i \in I_j} w_ih_i} 
     =-\frac{\sum_{i \in I_j^1} g_i+h_iv_j^*}{\sum_{i \in I_j^1} h_i}
\end{aligned}
$$
where $I_j^0 = \{i|q(\mathbf{X}_i) = j,w_i=0\}$ is the control instance set and $I_j^1 = \{i|q(\mathbf{X}_i) = j,w_i=1\}$ is the treated instance set of leaf $j$. 
It is obvious that, after obtaining $v^*$ from the control group, $u^*$ is only related to the treated samples. 
This innovation simplifies the original solution process to sequentially solving two quadratic equations in one variable. It also enables the CausalGBM algorithm to scale to multiple treatment scenarios with minimal additional computational resources, as $u^*$ is computed based on the samples of the corresponding group independently of other groups.
We then calculate the corresponding optimal loss by:
$$
\begin{aligned}
\tilde{\mathcal{L}}_{global}^{(t)}(q)=\sum_{j=1}^T [
(\sum_{i \in I_j} g_i)v_j^* +
\frac{1}{2}(\sum_{i \in I_j}h_i)(v_j^*)^2
-
\frac{(\sum_{i \in I_j^1} g_i+h_iv_j^*)^2}{2\sum_{i \in I_j^1}h_i}
]
\end{aligned}
$$
Note that the value is obtained by computing all instances on leaf $I_j$ to ensure the global loss is optimized under this approximation method. 

\subsection{Greedy Algorithm for Tree Construction}
Enumerating all possible tree structures to find the minimum loss is infeasible due to the combinatorial explosion. 
Instead, we employ a greedy algorithm that recursively bifurcates nodes, starting with a single parent node. 
We define the loss of a leaf as:
$$
\begin{aligned}
\tilde{\mathcal{L}}^{(t)}_{leaf}=(\sum_{i \in I_{leaf}} g_i)f_j^* +
\frac{1}{2}(&\sum_{i \in I_{leaf}}h_i)(f_j^*)^2 
-
\frac{(\sum_{i \in I_{leaf}^1} g_i+h_if_j^*)^2}{2\sum_{i \in I_{leaf}^1}h_i}
\end{aligned}
$$
Assume that $I_L$ and $I_R$ are the instance sets of left and right nodes after the split. 
Letting $I = I_L \cup I_R$, then the loss reduction after the split is given by:
$$
\tilde{\mathcal{L}}_{split}=\tilde{\mathcal{L}}^{(t)}_I-(\tilde{\mathcal{L}}^{(t)}_L + \tilde{\mathcal{L}}^{(t)}_R)
$$
The above function will be used to evaluate the candidate split points. 
Compared with the gbm algorithm, CausalGBM redefines the computation formulas for weights and evaluation functions in uplift modeling problems. 
As for the construction of the tree, we follow the computational framework of lightgbm but have adjusted the calculation methods pertaining to weights and evaluations.

%% file: content/experimental_evaluation.tex
\section{Experiments}
In this section we present an experimental evaluation of the two proposed algorithms and compare their performance with performance of the base models and bagging. 
To comprehensively evaluate our proposed methods, extensive experiments were conducted on three large-scale real-world datasets\cite{hillstrom2008minethatdata,diemert2018large} and a synthetic dataset\cite{causalML}. 
A summary of these datasets is given in Table \ref{table1}. 

\begin{table*}[htbp]
	\vspace{-0.5cm}
	\caption{The basic statistics of datasets used in the paper.}
	\label{table1}
	\centering
	\begin{tabular}{ccccc} 
		\hline
		Metrics                         & CRITEO & HILLSTROM & VOUCHER & SYNTHETIC$_{m}$   \\ 
		\hline\hline
		Size                         & 1,000,000 & 42,693    & 371,730 & 200,000  \\
		Features                      & 12        & 8         & 2076 & $m$      \\
		\hline
		Avg. Label                   & 0.047      & 0.129      & 0.356 & 0.600     \\
		Treatment Ratio                & 0.85       & 0.50       & 0.85  & 0.50     \\
		\hline
		Relative Uplift (\%)      & 26.7      & 42.6      & 2.0  & 50.0     \\
		\hline
	\end{tabular}
	\centering
	\vspace{-1.0cm}
\end{table*}

\subsubsection{\textbf{Evaluation Protocols}}
We perform 10-fold cross-validation and use Qini coefficient \cite{diemert2018large,gutierrez2017causal,2021A} (normalized by prefect Qini score) for evaluation, and we perform a grid search for hyperparameters to search for an optimal parameter set that achieved the best performance on the validation dataset, which consisted of 25\% of the training dataset in each fold. 
\subsection{Overall Performance Comparison}
\begin{table*}[htbp]
	\small
	\vspace{-0.3cm}
	\caption{Model performance evaluated by Qini coefficient on four datasets with corresponding mean and standard error. 
	"S-", "T-", "TO-" and "URF-" stands for instantiations of single-model\cite{athey2015machine}, two-model\cite{radcliffe2007using}, transformed outcome\cite{jaskowski2012uplift} and uplift random forest\cite{guelman2012random}, respectively. 
	The base learner for the methods in the first part of the table is Lightgbm. 
	For methods of "URF-", we select four splitting criteria based on KL divergence, $\chi^2$ divergence, Euclidean and the difference of uplift (DDP) between the two leaves for decision trees.
	}
	\label{metrics}
	\centering
	\begin{tabular}{lcccc} 
		\hline
		\multirow{2}{*}{Model} & HILLSTROM & CRITEO & VOUCHER & SYNTHETIC$_{100}$ \\
		& \multicolumn{4}{c}{Qini Coefficient ($mean \pm s.e.$)} \\ 
		\hline\hline
		S-LGB & $ 0.0616 \pm 0.018 $ & $ 0.0933 \pm 0.016 $ & $ 0.0032 \pm 0.005 $ & $ 0.1812 \pm 0.003 $ \\
		T-LGB & $ 0.0567 \pm 0.018 $ & $ 0.0900 \pm 0.018 $ & $ 0.0014 \pm 0.005 $ & $ 0.1831 \pm 0.002 $ \\
		TO-LGB & $ 0.0377 \pm 0.020 $ & $ 0.0941 \pm 0.020 $ & $ 0.0048 \pm 0.006 $ & $ 0.1832 \pm 0.004 $ \\
		X-Learner\cite{kunzel2019metalearners} & $ 0.0619 \pm 0.015 $ & $ 0.0929 \pm 0.025 $ & $ 0.0029 \pm 0.007 $ & $ 0.1824 \pm 0.003 $ \\
		R-Learner\cite{nie2021quasi} & $ 0.0621 \pm 0.021 $ & $ 0.0936 \pm 0.020 $ & $ 0.0033 \pm 0.007 $ & $ 0.1829 \pm 0.003 $ \\
		\hline
		TARNet\cite{shalit2017estimating} & $ \underline{0.0636 \pm 0.020} $ & $ 0.0935 \pm 0.011 $ & $ 0.0045 \pm 0.008 $          & $ 0.1803 \pm 0.005 $ \\
		CFRNet\cite{shalit2017estimating} & $ 0.0635 \pm 0.022 $ & $ 0.0909 \pm 0.017 $ & $ 0.0042 \pm 0.008 $ & $ 0.1829 \pm 0.004 $ \\
		\hline
		CForest\cite{wager2018estimation} & $ 0.0617 \pm 0.014 $ & $ 0.0933 \pm 0.011 $ & $ 0.0055 \pm 0.004 $ & $ 0.1395 \pm 0.004 $ \\
		UB-RF\cite{rafla2023parameter} & $ 0.0595 \pm 0.013 $ & $ \underline{0.0959 \pm 0.019} $ & $ 0.0081 \pm 0.009 $ & $ 0.1775 \pm 0.005 $ \\
		URF-Chi & $ 0.0623 \pm 0.017 $ & $ 0.0925 \pm 0.013 $ & $ 0.0062 \pm 0.007 $ & $ 0.1003 \pm 0.005 $ \\
		URF-ED & $ 0.0613 \pm 0.018 $ & $ 0.0942 \pm 0.014 $ & $ 0.0070 \pm 0.006 $ & $ 0.1657 \pm 0.006 $ \\
		URF-KL & $ 0.0605 \pm 0.016 $ & $ 0.0926 \pm 0.012 $ & $ 0.0060 \pm 0.006 $ & $ 0.1457 \pm 0.004 $ \\
		URF-DDP & $ 0.0599 \pm 0.016 $ & $ 0.0938 \pm 0.014 $ & $ 0.0072 \pm 0.006 $ & $ 0.1661 \pm 0.005 $ \\
		\hline
		TDDP & $ 0.0576 \pm 0.012 $ & $ 0.0884 \pm 0.018 $ & $ \underline{0.0088 \pm 0.006} $ & $ \underline{0.1836 \pm 0.005} $ \\
		CausalGBM & $ \boldsymbol{0.0643 \pm 0.025} $ & $ \boldsymbol{0.0971 \pm 0.014} $ & $ \boldsymbol{0.0108 \pm 0.004} $ & $ \boldsymbol{0.1863 \pm 0.004} $ \\
		\hline
	\end{tabular}
	\vspace{-0.75cm}
	\centering
\end{table*}

To verify that our proposed methods can make the uplift prediction model more accurate, we compare TDDP and CausalGBM with different types of baselines and show their prediction performance on four large-scale datasets in Table \ref{metrics}. Here, we summarize key observations and insights as follows:

\textbf{CausalGBM's superior performance}: Our proposed model CausalGBM outperforms all different baseline methods across all datasets. Specifically, it achieves relative performance gains of 1.1\%, 1.3\%, 22.7\%, and 1.5\% on four datasets, respectively, comparing to the best baseline. Further, Qini reflects the model's ability to give high predicted probabilities of persuasion to those who are actually more likely to be persuaded, and the improvement implies that our proposed model more accurately finds the target population for which the treatment is effective.

\textbf{CausalGBM's robustness across different scenarios}:Comparing results across four datasets of varying scales, many baseline models lack robustness. For instance, URF-based methods excel on HILLSTROM and CRITEO but perform poorly on SYNTHETIC$_{100}$, likely due to the smaller feature dimensions of the first two datasets. In contrast, CausalGBM consistently achieves the best performance across all datasets, demonstrating its robustness.

On the VOUCHER dataset, deep learning and some meta-learning methods, which focus on fitting potential outcomes, are weaker than tree models that emphasize causal effect heterogeneity. This dataset has the weakest causal effect significance, challenging methods that only consider potential outcomes, resulting in predicted causal effects close to zero. URF-based methods still perform well by minimizing heterogeneous differences.
CausalGBM, trained to compute both potential outcomes and causal effects simultaneously, remains robust on the weak causal effect dataset compared to methods driven solely by potential outcomes. However, on the HILLSTROM dataset, CausalGBM exhibited high variance, which was not observed on other datasets. This indicates that the algorithm shows some performance fluctuations on smaller-scale data.

\textbf{An analysis of the volatility for TDDP on different datasets}: 
The difference between TDDP and URF-DDP lies in the ensemble learning method. 
Comparing the performances of the two methods on four datasets, TDDP performs better on the dataset with high-dimensional features, while URF-DDP performs better on the dataset consisting of low-dimensional features. 
A plausible reason is that TDDP overfits the data in datasets with low-dimensional features, meanwhile, URF-DDP does not fit the data in high-dimensional datasets adequately.

\subsection{Analysis of Ensemble Method}
We will study how the ensemble learning method contributes to the predictive performance of the model in this section. 
In order to better visualize the experimental findings and prevent the derivation of conclusions from misleading information, we select the synthetic dataset and divide 50\% as the training set and the remaining part as the test. 
\begin{figure}[h]
	\centering
	\includegraphics[width=0.7\linewidth,height=0.55\linewidth]{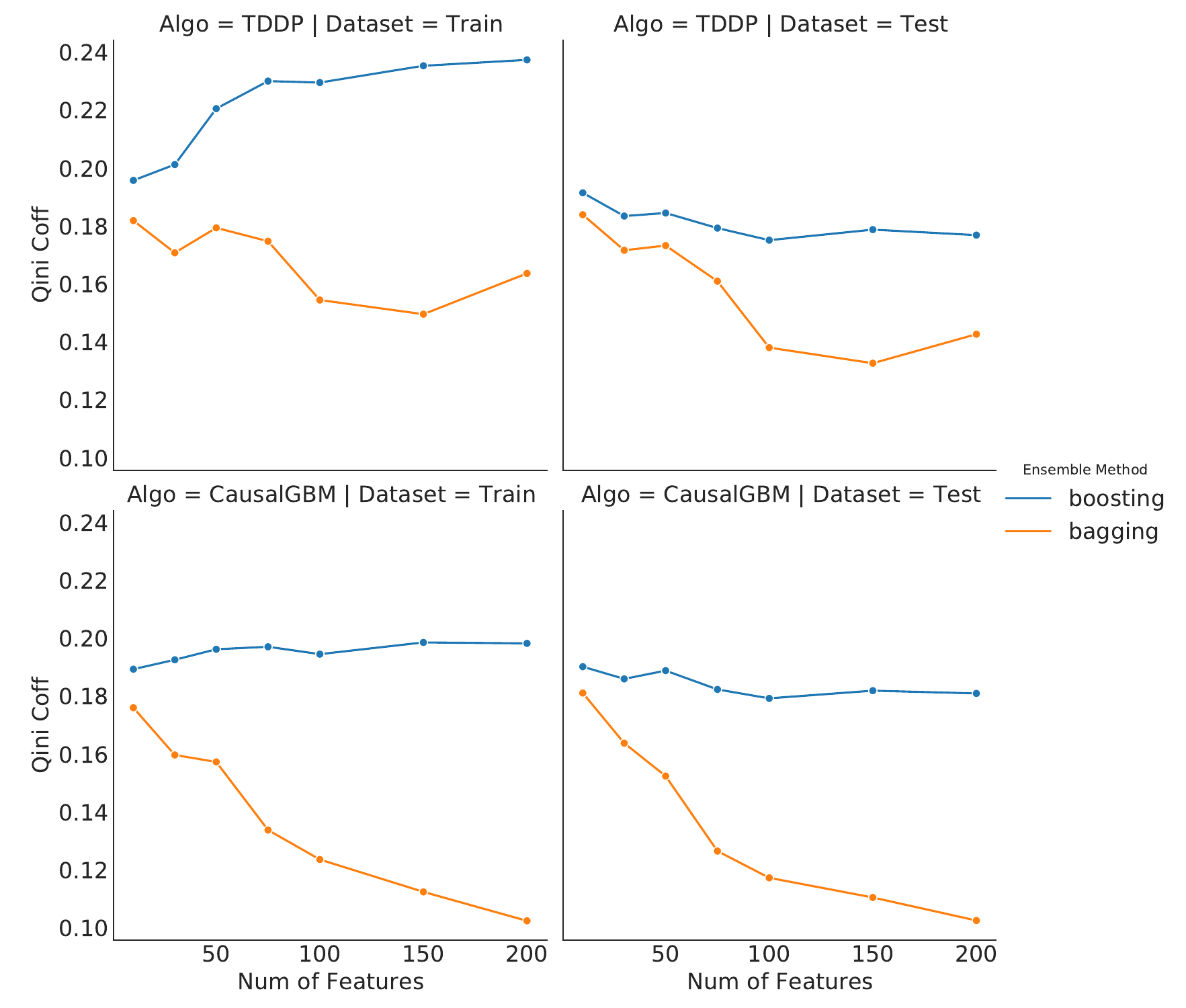}
	\caption{The result on different ensemble methods. 
	The upper and lower parts are the results of TDDP and CasualGBM respectively, while the left and right parts represent the training and testing datasets. 
	Two ensemble methods are distinguished by color.
	}
\end{figure}

We compare the prediction results of TDDP and CausalGBM with an ablation version using bagging instead of boosting. 
In this version, the gradient is computed only once before training and remains constant. 
We find that boosted trees significantly enhances the model's ability to fit the training data compared to bagging, especially as feature dimensionality increases. 
This improvement is also evident in the test dataset, indicating that boosting is particularly effective for high-dimensional datasets. 
On low-dimensional datasets, the difference between the methods is minimal. 
Additionally, the boosting version of TDDP tends to overfit with increasing feature size, leading to weaker generalization compared to CausalGBM.

In light of this finding, we suggest that tree models, which focus on the heterogeneity of local causal effects, require regularization methods to avoid overfitting, compared to the GBM approach that optimizes the global loss function.

%% file: content/conclusion.tex
\section{Conclusion}
In this paper, we formulate two novel boosting methods for the uplift modeling problem.
The first algorithm we propose follows the idea of maximizing the heterogeneity of causal effects. 
In contrast, the second algorithm we proposed, CausalGBM, fits both potential outcomes and causal effects by optimizing the loss function. 
We demonstrate that our proposed techniques outperform the baseline model on large-scale real datasets, where the CausalGBM algorithm shows excellent robustness, while the TDDP algorithm needs to blend in some regularization methods to prevent the model from overfitting the training data.